\pdfoutput=1

\documentclass[11pt]{article}

\usepackage{EACL2023}

\usepackage{times}
\usepackage{latexsym}

\usepackage[T1]{fontenc}

\usepackage[utf8]{inputenc}

\usepackage{microtype}

\usepackage{inconsolata}

\usepackage{mathtools}
\usepackage{comment}
\usepackage{booktabs}
\usepackage{multirow}
\usepackage{xcolor}
\usepackage[group-separator={,}]{siunitx}

\title{On the Role of Reviewer Expertise in Temporal Review \\Helpfulness Prediction}

\author{Mir Tafseer Nayeem \\
  University of Alberta \\
  \texttt{mnayeem@ualberta.ca} \\\And
  Davood Rafiei \\
  University of Alberta \\
  \texttt{drafiei@ualberta.ca} \\}

\begin{document}
\maketitle

\begin{abstract}

Helpful reviews have been essential for the success of e-commerce services, as they help customers make quick purchase decisions and benefit the merchants in their sales. While many reviews are informative, others provide little value and may contain spam, excessive appraisal, or unexpected biases. With the large volume of reviews and their uneven quality, the problem of detecting helpful reviews has drawn much attention lately. Existing methods for identifying helpful reviews primarily focus on review text and ignore the two key factors of (1) \textbf{who} post the reviews and (2) \textbf{when} the reviews are posted. Moreover, the helpfulness votes suffer from scarcity for less popular products and recently submitted (a.k.a., cold-start) reviews. To address these challenges, we introduce a dataset and develop a model that integrates the reviewer’s expertise, derived from the past review history of the reviewers, and the temporal dynamics of the reviews to automatically assess review helpfulness. We conduct experiments on our dataset to demonstrate the effectiveness of incorporating these factors and report improved results compared to several well-established baselines.

\end{abstract}

\section{Introduction}

\begin{figure}[t]
    \centering
    \includegraphics[scale = 0.66]{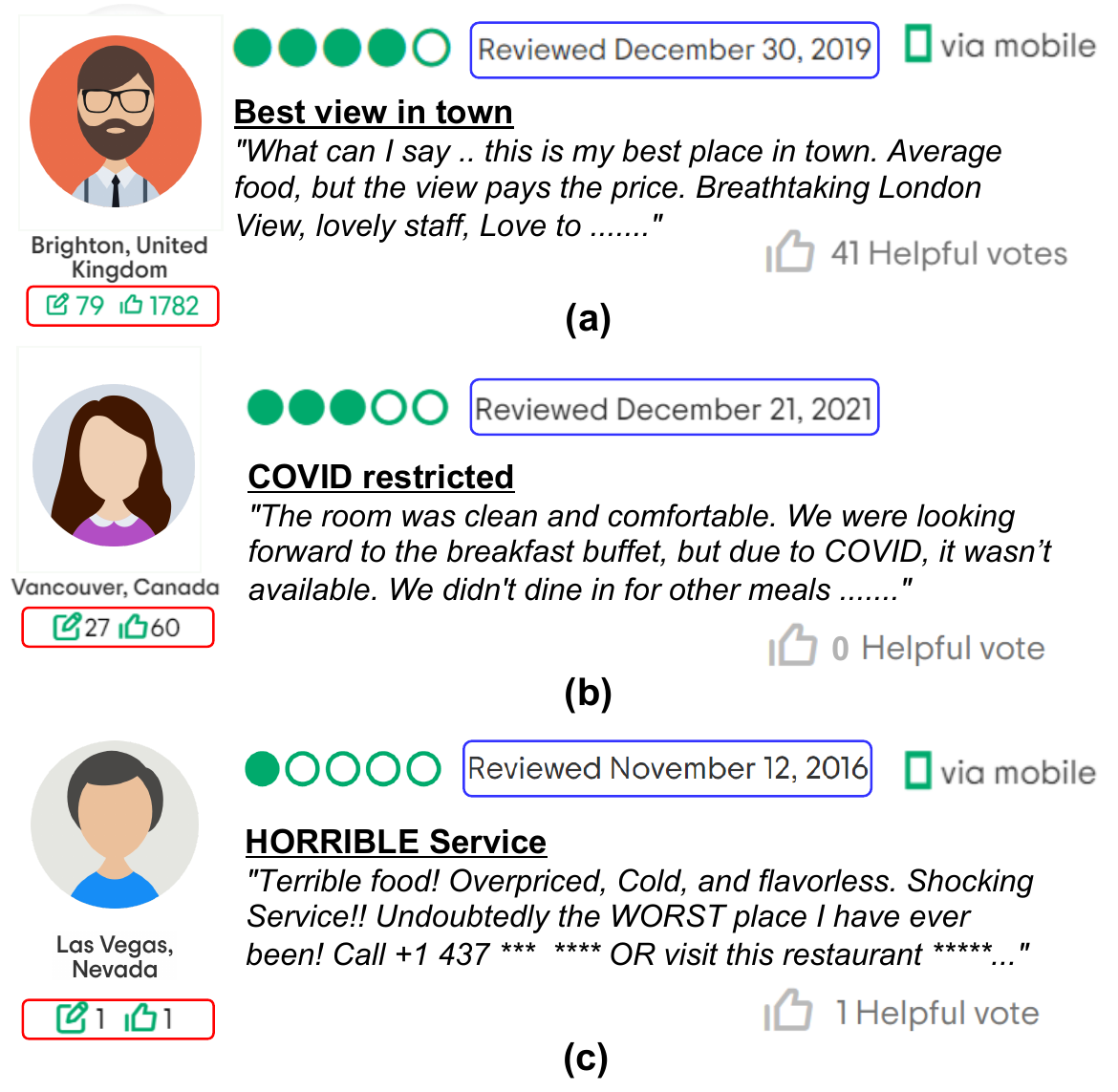}
    \caption{A snapshot of three reviews with the reviewers' history information: Review \textit{a} has accumulated more helpful votes but is posted almost two years before Review \textit{b}; on the other hand, Review \textit{b} (a.k.a., cold-start review) contains time-sensitive information, describing the current conditions and Review \textit{c} is likely a spam review. Photos of the reviewers are replaced with avatars for privacy reasons.}
    \label{fig:expl1}
\end{figure}

Many customers rely on online reviews from non-professionals, on daily basis, to decide what products to buy (e.g., \textit{Amazon}), what hotels to stay at (e.g., \textit{TripAdvisor}), what restaurants to eat (e.g., \textit{Yelp}) and even what books to read (e.g., \textit{Goodreads}). A recent survey of Bizrate Insights reward members found that approximately \emph{98\%} of online shoppers research a vendor via online reviews before making a purchase decision \cite{kats-article}. Since the reviews are expected to describe the actual experiences and opinions of users, they can provide a reliable source of reference, improving other customers’ confidence, comfort, and the overall shopping experience \citep{foo2017consistency, gamzu-etal-2021-identifying}. However, despite their tremendous benefits, online reviews are often of mixed qualities. While many reviews are informative, others provide little value and may contain excessive appraisal or spam (see Figure~\ref{fig:expl1}-c). There are multiple factors that affect the quality of a review, including the reviewers’ life experience, educational background, and the motive for writing the review \citep{vu42587}, and these factors are not usually explicit in the review text. All these pose challenges for customers who are less experienced in a subject area and need the reviews the most, simply because there is less incentive for more experienced users to use the reviews.
Moreover, customers usually have limited patience for reading reviews -- most customers read less than \num{10} reviews before making a purchase decision about an item \citep{murphy-article}. The large volume of reviews and their unpredictable quality and the limited customer patience demand better review utilization strategies to manage the information overload.

One standard method to identify more informative reviews is to ask for feedback from customers or site visitors who read them. By asking, \emph{“Was this review helpful to you?,”} or \emph{“Did you find this review helpful?”} at the end of each review, online platforms can crowdsource helpfulness votes from other customers. As a result, user reviews that gain the most helpful votes are shown first to the potential buyers to make the decision easier. However, the voting data suffers from scarcity \citep{10.1145/1772690.1772781} since only a tiny proportion of customers are willing to cast helpfulness votes. The scarcity is even more severe in reviews of less popular products and more recently submitted reviews (a.k.a., cold-start reviews) \citep{4781139}, despite the fact that more recent reviews may in fact contain more relevant and time-sensitive information (e.g., \emph{"New COVID Restrictions"} or \emph{"Dirty Pool Area"}) as shown in Figure~\ref{fig:expl1}-b but no helpfulness vote.

In this paper, we study the confluence of the reviewing history of reviewers and the review text for helpfulness identification. First, we observe that people who post more reviews and earn more helpful votes are more likely to be better reviewers. Second, trustworthy reviewers (e.g., Figure~\ref{fig:expl1}-a) are less likely to be posting fake or biased reviews, and their reviews are more likely to earn more helpful votes; otherwise, they will be ruining their reputation.
Third, those who have been to more hotels or restaurants across different cities have a better basis for comparison and writing critical reviews. 
To the best of our knowledge, existing works only focus on review content and neglect the reviewers and their reviewing history. Integrating the review text with the reviewing history of the reviewers is the problem studied in this paper. 

Our main contributions are summarized as follows:

\begin{itemize}
  \item We introduce a new dataset with both review text and reviewer's history, to highlight the importance of integrating the two sources for review helpfulness.
  
  \item We propose a model incorporating the reviewer's expertise and temporal information of reviews in helpfulness prediction. 
  
  \item We present a detailed case-study to interpret the model behavior and highlight potential directions to be addressed in the future. 
 
\end{itemize}
\section{Related work}

More traditional approaches on review helpfulness prediction focus solely on the text of reviews, and some consider both text and images to guide the prediction. In general, the task can be addressed using a predictive model based on hand-crafted features such as \texttt{structural} \citep{susan2010makes, xiong-litman-2014-empirical}, \texttt{lexical} \citep{10.5555/1610075.1610135, xiong-litman-2011-automatically}, \texttt{syntactic} \citep{10.5555/1610075.1610135}, \texttt{emotional} \citep{martin2014prediction}, \texttt{semantic} \citep{yang2015semantic}, and \texttt{arguments} \citep{liu-etal-2017-using} from the review text. These features may be fed into a conventional classifier such as SVM, Random Forest, or gradient boosting to identify helpful reviews. These methods heavily rely on manual feature engineering, which is labor-intensive and time-consuming. 

Inspired by the remarkable progress of deep neural networks, more recent studies make use of deep neural models, which can learn both intrinsic and extrinsic features given labeled data. \citet{chen-etal-2018-cross} uses a text-based CNN model to automatically capture the character-level, word-level, and topic-level features for helpfulness prediction. \citet{10.5555/3382225.3382298} uses an end-to-end multi-task neural architecture with the help of an auxiliary task, such as rating regression, to boost the performance of the review helpfulness identification. \citet{liu-etal-2021-multi-perspective} and \citet{han-etal-2022-sancl} use both text and images to guide the review helpfulness prediction. Since the image field is usually optional in reviews, a large volume of reviews contain only text, for which these multimodal models would produce inconsistent results.
\section{Review Helpfulness Prediction}

\subsection{Dataset}

To the best of our knowledge, there is no human-annotated dataset that is publicly available for the task of review helpfulness prediction with the reviewers' attributes and review date. Therefore, we build our dataset by scraping reviews from \texttt{TripAdvisor}\footnote{\url{https://www.tripadvisor.com}}. Out of \num{225664} reviews retrieved, close to one third have no helpful votes. We filter such reviews, and this reduces the number of reviews to \num{161541}. Table~\ref{tab:data-stats} presents the summary of our dataset with train, validation, and test splits\footnote{We present our dataset construction details in Section \ref{sec:app-dataset} of the Appendix.}. 
Following \citep{liu-etal-2021-multi-perspective}, we leverage a logarithmic scale ($\lfloor{\log_2 n_{\rm{votes}}}\rfloor$) to categorize the reviews based on the number of votes received. Specifically, we map the number of votes into five intervals (i.e., [1,2), [2, 4), [4, 8), [8, 16), [16, $\infty$)), each corresponding to a helpfulness score $Y$ $\in$ $\{1, 2, 3, 4, 5\}$, where the higher the score, the more helpful the review. 

\begin{table}[t]
\centering
\begin{tabular}{lccc}
\hline
\multicolumn{1}{c}{\textbf{}} & \multicolumn{1}{l}{\textbf{Train}} & \multicolumn{1}{l}{\textbf{Valid}} & \multicolumn{1}{l}{\textbf{Test}} \\ \hline \hline
Total \#Samples               & 145,381                            & 8,080                              & 8,080                             \\
Avg. \#Sentences              & 7.82                               & 7.80                                & 7.81                              \\
Avg. \#Words                  & 152.37                             & 152.25                             & 148.90                             \\ \hline
\end{tabular}

\caption{Our dataset statistics.}
\label{tab:data-stats}
\end{table}

\subsection{Proposed Model}

\textbf{R}eview \textbf{H}elpfulness \textbf{P}rediction (\textbf{RHP}) can be modeled as a supervised machine learning task where the input contains information about the reviews ($\mathcal{R}$) and the reviewers ($\mathcal{U}$). 
Let $\mathcal{R}_i = ([s_1, \ldots , s_N], \  t_i)$ denote a review posted at time $t_i$ with sentences $s_1, \ldots , s_N$, and $\mathcal{U}_i = (n_i, \ m_i)$ denote a reviewer who posts $n_i$ reviews and earns a total of $m_i$ helpful votes.
We formulate the review helpfulness prediction as a multi-class classification where we seek to find a model $f$ that minimizes the loss function $\mathcal{L}$, i.e.
\begin{equation}
 min_{\theta} \mathcal{L} \ (f(\theta, \mathcal{R}, \mathcal{U}), \ Y), 
\end{equation}
where $Y$ is the ground-truth, $\theta$ is the model parameter and the output of the model is a helpfulness class $\hat{Y}$ $\in$ $\{1, 2, 3, 4, 5\}$. The learning task is to find the best parameter that minimizes the above equation. 

We encode the review sentences using \texttt{BERT} \cite{devlin-etal-2019-bert, xu-etal-2019-bert}.
We concatenate the review sentences together while inserting a \texttt{[CLS]} token at the start and a \texttt{[SEP]} token at the end. If $\mathbf{h}^\texttt{[CLS]}$ denotes the embedding vector of the special \texttt{[CLS]} token and $\mathbf{h}^{(i)}$ denotes the embedding vector of the $i$-th token, we extract the last hidden state of  $\mathbf{h}_{l}^\texttt{[CLS]}$ to represent the review sentences and apply a linear transformation to get a final contextualized representation $x_h \in {\rm I\!R}^K$, where $\Theta$ is a non-linear activation function.
\begin{equation}
\begin{multlined}
[\mathbf{h}^\texttt{[CLS]}, \mathbf{h}^{(1)}, \mathbf{h}^{(2)}, \dots] =  \textbf{BERT}(\texttt{[CLS]} \\ \ s_1, \dots , s_N \ \texttt{[SEP]}),
\end{multlined}
\end{equation}
\begin{equation}
 x_h = \Theta \ ( \textbf{MLP} \ (\mathbf{h}_{l}^\texttt{[CLS]})).
\end{equation}
Generally, users who post more reviews and earn more helpful votes are likely to be better reviewers. Such users may have been to more hotels and restaurants across the globe and have a better basis for comparison. We define the term \textit{reviewer expertise} as the mean number of helpful votes received per review, written as $e_s = m/n$ for a reviewer who posts $m$ reviews and earns $n$ overall helpfulness votes.
We use a linear layer to get a weighted representation of the expertise score ($h_s$). 
%
%
\begin{equation}
 h_s = \textbf{MLP} \ (e_s)
 \label{eq:hs}
\end{equation}

Previous approaches for this task fail to consider the temporal nature of the reviews. Older reviews are more likely to accumulate more helpfulness votes than newer reviews but are not necessarily the most relevant describing the current conditions (e.g., \emph{new COVID restrictions}). One-time problems such as broken bathrooms and dirty pool area are likely to be addressed and to be less relevant. Let $t_d$ be the relative age of a review in days, for example, as of the day the reviews are scraped.
We use a linear layer to get a weighted representation of the relative review age.


\begin{equation}
 h_t = \textbf{MLP} \ (t_d).
 \label{eq:ht}
\end{equation}

It should be noted that both the review age and the reviewer expertise are normalized to a fixed range $[a,b]$ 
before being used in the linear layers in Equations~\ref{eq:hs} and \ref{eq:ht}.
If $\mathcal{X}$ denotes a set of scores (e.g., reviewers expertise score),
 a score $x_i \in \mathcal{X}$ is normalized into $z_i$ as follows:
\begin{equation}
z_i = (b-a)\frac{x_i - min(\mathcal{X})}{max(\mathcal{X}) - min(\mathcal{X})} + a
\end{equation}
%
%
%
In our case, both review age and reviewer expertise are scaled into the interval $[0,1]$.

We concatenate the textual representation ($x_h$), expertise representation ($h_s$), and temporal representation ($h_t$ ) to get a final embedding 
\begin{equation}
o_{final} = h_s \oplus x_h \oplus h_t,
\end{equation}
where $\oplus$ is a concatenation operator.
The final helpfulness prediction layer feeds $o_{final}$ into a linear layer and use softmax activation to get the final predicted helpfulness class $\hat{Y}$.
\begin{equation}
\hat{Y} =  \textbf{softmax} \ (W_r \cdot o_{final} + b_r),
\end{equation}
\begin{equation}
\mathcal{L} = \mathcal{L}_{CE} \ (\hat{Y}, Y)
\end{equation}
where $W_r \in {\rm I\!R}^{K \times K}$ and $b_r \in {\rm I\!R}^K$ denote the projection parameter and a bias term respectively. We use the cross-entropy loss function $\mathcal{L}_{CE}$ with respect to the ground truths  helpfulness class ($Y$).

\subsection{Experiments}

We evaluated the performance of the proposed model\footnote{Code, dataset, and model checkpoints: \url{https://github.com/tafseer-nayeem/RHP}} compared to well-established baselines. We compare our system with \texttt{ARH} \citep{10.5555/1610075.1610135}, \texttt{UGR + BGR} \citep{xiong-litman-2011-automatically}, \texttt{TextCNN} \citep{chen-etal-2018-cross}, \texttt{MTNL} \citep{10.5555/3382225.3382298}, and \texttt{BERTHelp} \citep{10.1007/978-3-030-39442-4_21}. We didn’t perform any explicit preprocessing of the review text. We discuss the baseline systems, preprocessing, and hyperparameters used for our experiments in Appendix (Section \ref{sec:app-baselines} \& Section \ref{sec:app-preprocess-hyper}).

\begin{table}[t]
\centering
\small
\begin{tabular}{cccc}
\hline
\textbf{Baseline Models}                       & \textbf{Acc. ($\uparrow$)} & \textbf{MAE ($\downarrow$)} & \textbf{MSE ($\downarrow$)} \\ \hline \hline
\textbf{ARH}    & 58.73         & 0.476           & 0.619           \\
\textbf{UGR + BGR}   & 62.76             & 0.464            & 0.674            \\
\textbf{TextCNN}   & 62.82             & 0.444            & 0.608            \\
\textbf{MTNL}   & 62.77             & 0.458            & 0.653            \\ 
\textbf{BERTHelp}   & 63.03             & 0.432            & 0.591            \\  \hline
\textbf{Our Ablations}                         & \textbf{Acc. ($\uparrow$)} & \textbf{MAE ($\downarrow$)} & \textbf{MSE ($\downarrow$)} \\ \hline \hline
\multicolumn{1}{l}{\textbf{RHP (\textit{ours})}} & \textbf{65.18$^\dagger$} & \textbf{0.393$^\dagger$} & \textbf{0.491$^\dagger$} \\
\multicolumn{1}{l}{ \quad - \textit{w/o Expertise}}            & 63.87         & 0.421$^\dagger$        & 0.550$^\dagger$        \\
\multicolumn{1}{l}{ \quad - \textit{w/o Temporal}}             & 63.40         & 0.437$^\dagger$        & 0.592        \\
\multicolumn{1}{l}{ \quad - \textit{w/o Expertise + Temporal}} & 62.92         & 0.446        & 0.617        \\ \hline
\end{tabular}
\caption{ Performance compared to our baseline models and the result of our ablation study ($\uparrow$ indicates higher values for a better performance and $\downarrow$ indicates lower values for a better performance). $^\dagger$ reported results are statistically significant in paired t-test by taking \texttt{BERTHelp} \citep{10.1007/978-3-030-39442-4_21} as a reference with the confidence of 95\% ($p$-value < 0.05).}
\label{tab:results}
\end{table}

\subsubsection{Results}

As part of a detailed evaluation of our algorithm, we report our model’s performance compared with the baselines in terms of Accuracy (\textbf{Acc.}), Mean Average Error (\textbf{MAE}), and Mean Squared Error (\textbf{MSE}). As shown in Table~\ref{tab:results}, our final model outperforms the baselines in terms of all the metrics. Our ground-truth values consist of \num{5} classes which correspond to five helpfulness scores $\{1, 2, 3, 4, 5\}$, where the higher the score, the more helpful the review. To gain more insights into the performance of our prediction model, we also evaluate our algorithm in terms of \textbf{MAE} and \textbf{MSE}, which assess the fine-grained differences between the ground-truth and the predicted helpfulness scores. Our \textbf{RHP} model consistently outperforms the baselines with a good margin, which means when misclassified, our model predictions are very close to the actual helpfulness scores. We conduct detailed ablation studies to demonstrate the effects of different components of our \textbf{RHP} model by removing expertise (denoted as \textit{w/o Expertise}) and removing temporal information (denoted as \textit{w/o Temporal}). The ablation test results on our dataset are summarized in Table~\ref{tab:results}. We can observe that the temporal feature has the largest impact on the performance of our model, and the impact of expertise is also significant. This suggests that the reviewer’s expertise and temporal information of the reviews play a key role in review helpfulness prediction. Therefore, it is no surprise that combining all components achieves the best performance on our proposed dataset.

\subsubsection{Analysis}
\label{sec:analysis}

We also present a detailed analysis to provide more supportive evidence of our arguments. To this end, we randomly selected $m$ examples for each class of reviews considering helpfulness votes. Then, we extract Top $K$ (where $K$ = \num{5}) $n$-grams from each class of reviews to identify the most relevant keywords or topics in reviews to assess what aspects are most talked about the items (e.g., hotels or restaurants).

\paragraph{Preprocessing} Our preprocessing step includes tokenization, lemmatization, removal of stopwords, Part-Of-Speech (POS) tagging, and filtering punctuation marks. We use the NLTK\footnote{\url{https://www.nltk.org/}} to preprocess each sentence and obtain a more accurate representation of the information. Moreover, we also add \texttt{‘hotel’} and \texttt{‘restaurant’} in the stopwords list as they frequently occur in every review and are not meaningful in our context.

\paragraph{Extracting Candidate $n$-grams} We remove the sentiment words and emojis using \texttt{VADER}\footnote{\url{https://github.com/cjhutto/vaderSentiment}} \cite{Hutto_Gilbert_2014}, a \textit{"gold-standard"} sentiment lexicon especially attuned to microblog-like contexts. As the sentiment expressed in reviews are highly subjective, we are interested in extracting only the aspects or topics (e.g., \textit{room}, \textit{location}, \textit{customer service} etc.) for which the opinions are expressed. Therefore, we keep only the nouns\footnote{As adjectives and adverbs may contain sentiment towards aspects.} (with POS tags \texttt{`NN'} and \texttt{`NNS'}) for extracting the aspects or topics.

\paragraph{Ranking Candidate $n$-grams} We extract the unigrams and bigram collocations for each of the review classes. Then, we rank the unigrams by counting the frequency of occurrences and bigrams using likelihood ratios \cite{manning1999foundations} to obtain Top $K$. We present the Top \num{5} unigrams and bigrams in Table \ref{tab:further-analysis} grouped according to helpfulness classes and ordered by descending ranking scores.

\begin{table}[t]
\centering
\small
\begin{tabular}{clclc}
\hline 
\textbf{Helpfulness Class} &
   &
  \textbf{Unigram} &
   &
  \textbf{Bigram} \\ \hline \hline
 &
   &
  {\color[HTML]{009901} \textbf{`room'}} &
   &
  {\color[HTML]{009901} \textbf{`front desk'}} \\
 &
   &
  {\color[HTML]{009901} \textbf{`staff'}} &
   &
  {\color[HTML]{D72424} \textbf{`coffee maker'}} \\
 &
   &
  {\color[HTML]{D72424} \textbf{`location'}} &
   &
  `breakfast buffet' \\
 &
   &
  {\color[HTML]{009901} \textbf{`time'}} &
   &
  `sofa bed' \\
\multirow{-5}{*}{\textbf{\begin{tabular}[c]{@{}c@{}}Class \#1\\ Helpful Votes {[}1, 2)\end{tabular}}} &
   &
  {\color[HTML]{009901} \textbf{`service'}} &
   &
  `swim pool' \\ \hline \hline
 &
   &
  {\color[HTML]{009901} \textbf{`room'}} &
   &
  {\color[HTML]{009901} \textbf{`front desk'}} \\
 &
   &
  {\color[HTML]{009901} \textbf{`staff'}} &
   &
  `shampoo conditioner' \\
 &
   &
  {\color[HTML]{009901} \textbf{`service'}} &
   &
  {\color[HTML]{CE6301} \textbf{`customer service'}} \\
 &
   &
  {\color[HTML]{D72424} \textbf{`location'}} &
   &
  {\color[HTML]{3531FF} \textbf{`resort fee'}} \\
\multirow{-5}{*}{\textbf{\begin{tabular}[c]{@{}c@{}}Class \#2\\ Helpful Votes {[}2, 4)\end{tabular}}} &
   &
  {\color[HTML]{009901} \textbf{`time'}} &
   &
  `pool area' \\ \hline \hline
 &
   &
  {\color[HTML]{009901} \textbf{`room'}} &
   &
  {\color[HTML]{009901} \textbf{`front desk'}} \\
 &
   &
  {\color[HTML]{009901} \textbf{`staff'}} &
   &
  {\color[HTML]{3531FF} \textbf{`resort fee'}} \\
 &
   &
  {\color[HTML]{009901} \textbf{`time'}} &
   &
  {\color[HTML]{CE6301} \textbf{`customer service'}} \\
 &
   &
  {\color[HTML]{009901} \textbf{`service'}} &
   &
  {\color[HTML]{D72424} \textbf{`coffee maker'}} \\
\multirow{-5}{*}{\textbf{\begin{tabular}[c]{@{}c@{}}Class \#3\\ Helpful Votes {[}4, 8)\end{tabular}}} &
   &
  `view' &
   &
  `city view' \\ \hline \hline
 &
   &
  {\color[HTML]{009901} \textbf{`room'}} &
   &
  {\color[HTML]{009901} \textbf{`front desk'}} \\
 &
   &
  {\color[HTML]{009901} \textbf{`staff'}} &
   &
  {\color[HTML]{3531FF} \textbf{`resort fee'}} \\
 &
   &
  {\color[HTML]{009901} \textbf{`service'}} &
   &
  {\color[HTML]{CE6301} \textbf{`customer service'}} \\
 &
   &
  {\color[HTML]{009901} \textbf{`time'}} &
   &
  `minute walk' \\
\multirow{-5}{*}{\textbf{\begin{tabular}[c]{@{}c@{}}Class \#4\\ Helpful Votes {[}8, 16)\end{tabular}}} &
   &
  {\color[HTML]{D72424} \textbf{`pool'}} &
   &
  `life jacket' \\ \hline \hline
 &
   &
  {\color[HTML]{009901} \textbf{`room'}} &
   &
  {\color[HTML]{009901} \textbf{`front desk'}} \\
 &
   &
  {\color[HTML]{009901} \textbf{`time'}} &
   &
  {\color[HTML]{3531FF} \textbf{`resort fee'}} \\
 &
   &
  {\color[HTML]{009901} \textbf{`service'}} &
   &
  `bed bug' \\
 &
   &
  {\color[HTML]{009901} \textbf{`staff'}} &
   &
  `beach chair' \\
\multirow{-5}{*}{\textbf{\begin{tabular}[c]{@{}c@{}}Class \#5\\ Helpful Votes {[}16, $\infty$)\end{tabular}}} &
   &
  {\color[HTML]{D72424} \textbf{`pool'}} &
   &
  `cable car' \\ \hline \hline
\end{tabular}
\caption{Top \num{5} unigrams and bigrams extracted from five different classes of reviews divided according to helpfulness votes. For each column, {\color[HTML]{009901} \textbf{green}} color indicates the overlap with all \num{5} classes, whereas {\color[HTML]{3531FF} \textbf{blue}} for \num{4}, {\color[HTML]{CE6301} \textbf{orange}} for \num{3}, and {\color[HTML]{D72424} \textbf{red}} for \num{2} overlaps.}
\label{tab:further-analysis}
\end{table}

Table \ref{tab:further-analysis} shows a high overlap of $n$-grams among different classes of reviews, which further strengthen our argument that helpfulness does not entirely depend on the review text but rather the confluence of the review text, reviewing history of reviewers (\textit{\textbf{who} post the reviews}), review age (\textit{\textbf{when} the reviews are posted}). Generally, older reviews (i.e., review age) were present longer than the newer reviews in the platform and had more time to accumulate helpful votes.

\begin{figure}[t]
    \centering
    \includegraphics[scale = 0.61]{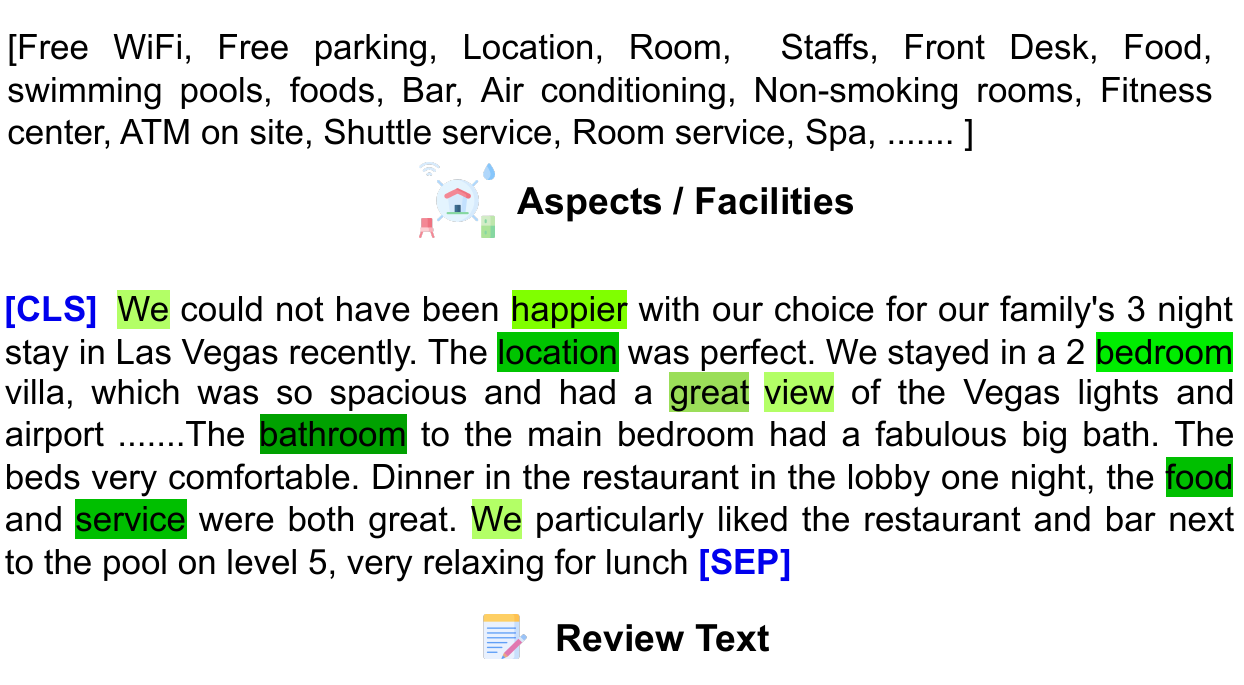}
    \caption{Top \num{10} ranked tokens of the \textbf{RHP} model shown in green colors with the color intensity indicating the importance of the tokens in the overall prediction.}
    \label{fig:case-study}
\end{figure}

\subsection{Case Study}

To gain more insights into the review helpfulness prediction task, we present a detailed case-study to interpret the model behavior and highlight the most important features of this task. Models are interpretable when humans can readily comprehend the reasoning behind model predictions and decisions made \citep{10.5555/3157096.3157352}. To this end, we randomly selected a sample with Helpfulness Class = \num{3} from our test set and used \texttt{Captum}\footnote{Captum (\url{https://captum.ai/}) is an open-source, extensible library for model interpretability that uses the integrated gradients method \citep{pmlr-v70-sundararajan17a}.} to interpret the words/tokens that contributed the most to the prediction. As can be seen in Figure~\ref{fig:case-study}, the top-ranked words are highly representative of the aspects/facilities listed on the restaurant page. We can conclude from this observation that users tend to look for specific aspects in reviews to find them helpful. We also notice that the use of personal pronouns (e.g., I, we, they, etc.), describing personal experiences, contributes to the helpfulness prediction. People often find reviews useful if it comes from others' experiences and personal pronouns are a good indicator of it.

\section{Conclusion and Future Work}

In this paper, we develop a model incorporating the reviewer’s expertise and temporal information in reviews to predict the helpfulness, especially for unreliable and cold-start reviews. 
Furthermore, we present a detailed analysis to interpret the model behavior and provide reasoning behind model predictions. For future work, we will look into the problem of personalized review helpfulness prediction to model the demographics and cultural differences of the reviewers.

\section*{Limitations}
Despite the effectiveness of incorporating the reviewer's history and temporal information of the reviews in helpfulness prediction, our current studies still have several limitations, which can pave the path for future research.

For simplicity, like existing works, we assume that all the users rate reviews unanimously. However, the diversity of demographics, age, and cultural background also affect how users give, receive, and understand the sentiments expressed in reviews. Users may focus on different review aspects based on their preferences (i.e., \emph{"5 stars, party every night"} \textbf{vs} \emph{"5 stars, always quiet and peaceful"}). It would be interesting to see how to incorporate personal preferences for the helpfulness prediction task.

Another limitation of our work is that we only worked with reviews written in English. As a result, we filter out the reviews written in other languages and notice code-switched reviews when the reviewers alternate between two or more languages in a single review. We aim to extend this work to support more languages.

\section*{Ethics Statement}

In our data scraping process, we took into account ethical considerations. We obtained data at an appropriate pace, avoiding any potential DDoS attacks. Additionally, we eliminated any Personal Identifying Information, such as names, telephone numbers, and email addresses, from the data set.
\section*{Acknowledgements}

We thank all the anonymous reviewers for their valuable feedback and constructive suggestions for improving this work. This research is supported by the Natural Sciences and Engineering Research Council of Canada (NSERC) and by a grant from Huawei. Mir Tafseer Nayeem is also supported by a Huawei Doctoral Scholarship.

\bibliography{eacl2023}
\bibliographystyle{acl_natbib}

\bigskip
\smallskip
\appendix

\section{Dataset Construction}
\label{sec:app-dataset}

Publicly available datasets which are mostly used for this task are \texttt{Amazon}\footnote{\url{http://jmcauley.ucsd.edu/data/amazon/index\_2014.html}} \cite{10.1145/2872427.2883037, 10.1145/2766462.2767755} and \texttt{Yelp}\footnote{\url{https://www.yelp.com/dataset}}. In Yelp dataset, the user votes are distributed among three categories such as \textit{“Useful”}, \textit{“Funny”} or \textit{“Cool”}, where \textit{“Useful”} voting feature was introduced much later than the other two categories. Therefore, many good reviews already in the dataset may not have been marked useful. On the other hand, the Amazon dataset does not contain the reviewers' reviewing history and helpfulness votes to evaluate our hypothesis studied in this paper. Moreover, for Amazon, the samples come from various categories such as Books, Electronics, Clothing, Beauty, Shoes and Jewelry, Grocery, Pet Supplies, etc – the total helpfulness votes for the reviewers are coming from different categories and it’s not explicit in the fields from Amazon website. Therefore, it’s hard to devise expertise because of domain diversity. 

We build our dataset by scraping reviews from \texttt{TripAdvisor}\footnote{\url{https://www.tripadvisor.com}}, a travel site that offers online hotel and restaurant reservations and a platform for sharing the travel experiences of users. We take reviews from January 1st, 2015 until January 1st, 2020, and extract only those written in English. For each review, we extract the review text, the total helpfulness votes and the posting time, and for each reviewer, we extract the number of reviews contributed and the cumulative helpfulness votes. The attributes we extracted are summarized as follows:

\bigskip

\begin{itemize}
\item \textbf{Reviews}
    \begin{itemize}
        \item Review Text
        \item Total Review Helpful Votes
        \item Review Posting Time
    \end{itemize}

\item \textbf{Reviewers}
    \begin{itemize}
        \item Total Number of Reviews Contributed
        \item Cumulative Helpful Votes
    \end{itemize}
\end{itemize}

\smallskip

\section{Baseline Systems}
\label{sec:app-baselines}

We compare our system performance with the following baselines.

\begin{itemize}
  \item \texttt{\textbf{ARH}} \citep{10.5555/1610075.1610135} \& \texttt{\textbf{UGR + BGR}} \citep{xiong-litman-2011-automatically} use machine learning-based methods with hand-crafted features such as \textit{structural}, \textit{lexical}, \textit{syntactic}, \textit{emotional}, \textit{semantic}, and \textit{meta-data} from the review text to address this task. These features are fed into conventional classifiers such as SVM, Random Forest, and gradient boosting to identify helpful reviews.
  
  \item \texttt{\textbf{TextCNN}} \citep{chen-etal-2018-cross} employs a text-based CNN model \cite{kim-2014-convolutional} to automatically capture the character-level, word-level, and topic-level features for helpfulness prediction.
  
  \item \texttt{\textbf{MTNL}} \citep{10.5555/3382225.3382298} utilizes end-to-end multi-task neural learning (MTNL) architecture for classifying helpful reviews. They take the help of an auxiliary task, such as rating regression, to boost the performance of the original task, which is review helpfulness identification.
  
  \item \texttt{\textbf{BERTHelp}} \citep{10.1007/978-3-030-39442-4_21} develop their helpfulness prediction model using pre-trained BERT \cite{devlin-etal-2019-bert}. They design a regression model using BERT-based features extracted from review texts, star rating, and product type information from Amazon product review dataset \cite{10.1145/2872427.2883037}.
 
\end{itemize}

\smallskip

\section{Preprocessing \& Hyperparameters}
\label{sec:app-preprocess-hyper}

\paragraph{Preprocessing} We didn’t perform any explicit preprocessing of the review text. Instead, we use BertTokenizer to avoid the out-of-vocabulary (\textbf{OOV}) problem, which uses \texttt{WordPiece} \cite{ wu2016google} for tokenizing the sentences into words or subwords. In addition, we add special tokens to the start (e.g., \texttt{[CLS]}) and end of each review text (e.g., \texttt{[SEP]}) and truncate all sentences to a single constant length (e.g., \num{512}).

\paragraph{Hyperparameters} We use Adam optimizer \citep{kingma2014adam} with a learning rate of $3 \times e^{-5}$ and a batch size of \num{32}. We use $\text{BERT}_{\text{BASE}}$ \citep{wolf-etal-2020-transformers} pre-trained model with a fixed vocabulary. We run the training for \num{5} epochs and check the improvement of validation (\textit{dev set}) loss to save the latest best model during training.

\end{document}